# Cotton Yield Prediction Using Random Forest

Alakananda Mitra [1, 2], Sahila Beegum [1, 3], David Fleisher [1], Vangimalla R. Reddy [1], Wenguang Sun [1, 3], Chittaranjan Ray [3], Dennis Timlin [1], Arindam Malakar [4]

[1] Adaptive Cropping System Laboratory, USDA-ARS, Beltsville, MD 20705, USA

[2] Nebraska Water Center, at the Institute of Agriculture and Natural Resources, University of Nebraska, Lincoln, NE 68588-6204, USA

[3] Nebraska Water Center, *part of* the Robert B. Daugherty Water for Food Global Institute, 2021 Transformation Drive, University of Nebraska, Lincoln, NE 68588-6204, USA

[4] Nebraska Water Center, *part of* the Robert B. Daugherty Water for Food Global Institute and School of Natural Resources, University of Nebraska, Lincoln, NE 68583-0844, USA

Corresponding author: amitra6@unl.edu

**Abstract:**

The cotton industry in the United States is committed to sustainable production practices that minimize water, land, and energy use while improving soil health and cotton output. Climate smart agricultural technologies are being developed to boost yields while decreasing operating expenses. Crop yield prediction, on the other hand, is difficult because of the complex and nonlinear impacts of cultivar, soil type, management, pest and disease, climate, and weather patterns on crops. To solve this issue, we employ machine learning (ML) to forecast production while considering climate change, soil diversity, cultivar, and inorganic nitrogen levels. From the 1980s to the 1990s, field data were gathered across the southern cotton belt of the United States. To capture the most current effects of climate change over the previous six years, a second data source was produced using the process-based crop model, GOSSYM. We concentrated our efforts on three distinct areas inside each of the three southern states: Texas, Mississippi, and Georgia. To simplify the amount of computations, accumulated heat units (AHU) for each set of experimental data were employed as an analogy to use time-series weather data. The Random Forest Regressor yielded a 97.75% accuracy rate, with a root mean square error of 55.05 kg/ha and an $R^2$ of around 0.98. These findings demonstrate how an ML technique may be developed and applied as a reliable and easy-to-use model to support the cotton climate-smart initiative.

## 1. Introduction:

Predicting crop yields is essential for addressing issues with food security in the face of global climate change. Policymakers and farmers alike stress the need for precise and timely yield forecast (Jeong, et al. 2016). Accurate yield estimates enable farmers to adopt suitable management measures and make better educated financial decisions [1]. The many factors at play, including crop-specific data, management techniques, shifting weather patterns and climate [2, 3], cultivars, soil types, pests, and diseases, make it difficult to estimate crop yields with accuracy. These variables have intricated and nonlinear effects on crops [4, 5, 6]. It makes a crop yield forecast model difficult to build.

The objective of this study is to accurately forecast cotton output by taking account of the impacts of climate change, particularly the high temperatures in the southern cotton belt of the United States. It is an example how smart agriculture [7] can assist in crop modeling. The article's remaining sections are arranged as follows: The method and the experimental verification are described in Section 2. In Section 3, the findings have been recorded, examined, and contrasted with the previous research. Section 4 concludes with a summary of the findings.

## 2. Methods:

Temperature is the main factor that influences the growth of cotton. So, among all the weather variables, only temperature was considered. But since weather variables are time series data, including them into crop models becomes more difficult and requires more dynamic computing. We transformed the time series temperature data into a scalar number, the accumulated heat unit (AHU), to simplify things. Eq.1 and 2 were used to calculate the AHU for the whole season.

$$DD_{60} = \frac{(°F_{max} + °F_{min})}{2} - 60 \qquad (1)$$

$$AHU = \sum_{n=1}^{N} DD_{60_n} \qquad (2)$$

Where $DD_{60}$ is the heat unit, $T_{max}$ and $T_{min}$ are the maximum and minimum temperature of a day, and N is the number of days in a cotton season. The other inputs are soil types, cultivars, and different amounts of N (from the fertilizers). Random Forest (RF) regressor was used as the ML model to predict from the input data. A basic RF regressor with 10 estimators was used.

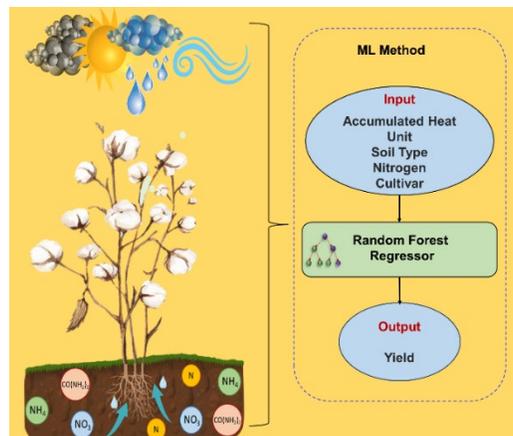

*Figure 1: Cotton Yield Prediction Overview.*

Initially, we started the work with field data. However, that dataset was not large enough to have a high performance from the Machine Learning (ML) algorithm, as ML algorithms require large dataset to perform better. To overcome this issue, we generated some synthetic data from the process-based cotton model GOSSYM [8, 9, 10]. Table.1 depicts the field dataset and Table. 2 describes the synthetic dataset details. Data preprocessing plays a major role in ML model building. Two types of data preprocessing techniques were used: outliers' removal and changing categorical data into numeric form.

Table 1 – Field Data Details

| Items | Details |
|---|---|
| Study States | California, Alabama, Missouri, Texas, Mississippi, Tennessee, New Mexico, |
| Number of Locations | 48 |
| Number of Study Years | 7 |
| Number of Soil Types | 10 |
| Number of Cultivars | 2 |
| Nitrogen Amount | 0-300 kg/ha |

Table 2 – Synthetic Data Details

| Study Location | Year | Cultivar | Soil | Nitrogen |
|---|---|---|---|---|
| Locations from Table 2 | 2017 - 2022 | 2 Upland Varieties | 3 types of Soil | 4 different values match the field data. |

The implementation details are as below.

- Keras [12] with the TensorFlow backend.
- Libraries used: pandas, numpy, matplotlib, sklearn etc.
- Trained on Intel Xeon Server with a 16-core CPU, 64 GB RAM, NVIDIA RTX A4000 GPU.
- Took a few minutes to train.
- A part of the dataset was kept held in the beginning and used as the test dataset. The train and valid dataset ratio was kept at 80:20.

3. Results

Three different performance metrics: RMSE, $R^2$, and accuracy were calculated to evaluate the model. A very high $R^2$ score of 0.98 was achieved. The RMSE was 55.05 kg/ha and the accuracy was 98.75%. Fig.

2 shows the RF regressor performance on the test dataset. Table 3 compares the work with other existing works.

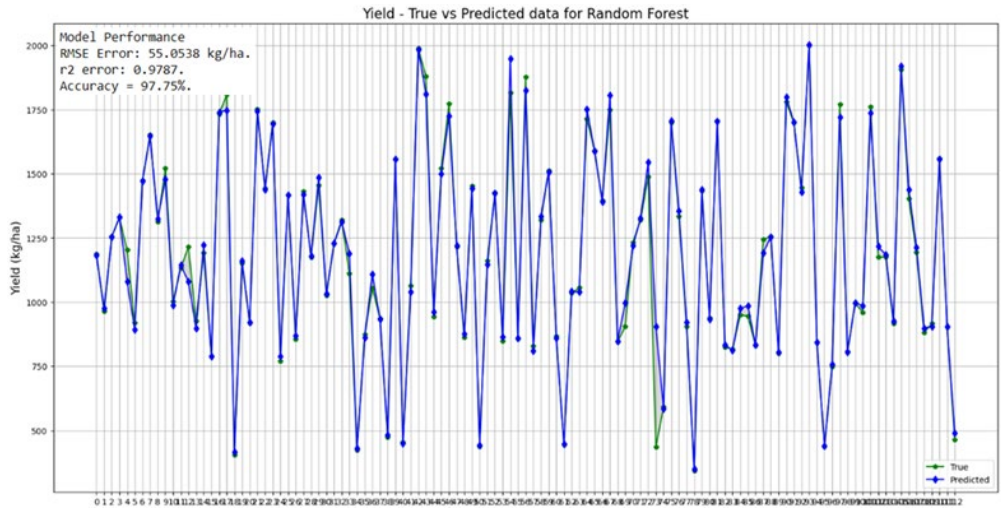

Figure 2: RF Regressor Performance on Test Dataset

Table 3- Comparative Analysis

| Crop | Method | Study Area + Study Period | RMSE (kg/ha) | R² Error | Remarks |
|---|---|---|---|---|---|
| Cotton | MODIS + RF | 5 locations in Maharashtra, India + Years: 2001-2017 | 62.77 | 0.69 | [13] |
| Cotton | Spatial-Temporal M.T.L. | A 48-ha location in west TX, USA +Years: 2001-2003 | 83.7 | - | [14] |
| Cotton | Spatial Temporal + RF + GBM | 2 locations in New South Wales, Australia + Years: 2014, 2016, 2017 | 170 (RF) 190(GBM) | 0.44 0.39 | [15] |
| **Cotton** | **AHU + RF** | **9 locations in USA+ Years: 2017-2022** | **55.05** | **0.98** | **Current Work** |

Table 7 presents a comparison of the regional and chronological diversity of our study, encompassing 9 sites and 6 recent years of data. Weather data from more recent years has been utilized to address the implications of climate change that have occurred recently. To prevent bias in weather data, the early years have not been purposefully included. Our approach is therefore far more reliable. However, more global locations, varied soil, cultivars, and nitrogen fluctuations must be added to the training data to create a more universal model (across the globe).

4. **Conclusions**

In this paper, using AHU data and an RF regressor, we were able to forecast the cotton output in a straightforward and very accurate manner. The work is unique in that it uses time-series weather data to create a scalar value, or AHU, without compromising model accuracy, thereby cutting down on computation. LightGBM regressor, an additional alternative machine learning method that we created, outperforms RF regressor in terms of performance. Large datasets are ideal for training machine learning and deep learning algorithms. Large publicly accessible datasets are not always accessible, which makes it difficult to use ML/DL-based techniques. When there is a lack of data, using synthetic data to train AI/ML models is a typical solution; nevertheless, it is not frequently employed in agriculture.

In addition to field data, this study employed produced data from a process-based cotton model. This work illustrated how ML/DL models in agriculture might be trained using synthetic data. Because of this study's excellent accuracy, we think that the gap between sophisticated technologies and the agricultural business will close and more research using synthetic data will be conducted in agriculture.

**Acknowledgment**

The authors are thankful to the Advanced Cropping Systems Laboratory, USDA-ARS for providing access to use the field dataset.

**References**

[1] J. Ansarifar, L. Wang and S. Archontoulis, "An interaction regression model for crop yield prediction," *Scientific Report,* vol. 11, no. 17754, 2021.

[2] D. Paudel, H. Boogaard, A. de Wit, S. Janssen, S. Osinga, C. Pylianidis and I. Athanasiadis, "Machine learning for large-scale crop yield forecasting," *Agricultural Systems,* vol. 187, p. 103016, 2021.

[3] R. Fischer, "Definitions and determination of crop yield, yield gaps, and of rates of change," *Field Crops Research,* pp. 9-18, 2015.

[4] D. B. Lobell, M. Bänziger, C. Magorokosho and B. Vivek, "Nonlinear heat effects on african maize as evidenced by historical yield trials," *Nature Climate Change,* vol. 1, no. 1, pp. 42-45, 2011.

[5] M. C. Broberg, P. Högy, Z. Feng and H. Pleijel, "Effects of elevated CO2 on wheat yield: Non-linear response and relation to site productivity," *Agronomy,* vol. 9, no. 5, p. 243, 2019.

[6] K. Lamsal, G. Paudyal and M. Saeed, "Model for assessing impact of salinity on soil water availability and crop yield," *Agricultural Water Management,* vol. 41, no. 1, pp. 57-70, 1999.

[7] A. Mitra, S. P. Mohanty and E. Kougianos, "Smart Agriculture--Demystified," in *IFIP International Internet of Things Conference*, 2023.

[8] D. N. Baker, J. D. Hesketh and W. G. Duncan, "Simulation of Growth and Yield in Cotton: I. Gross Photosynthesis, Respiration, and Growth 1," *Crop Science,* vol. 12, no. 4, pp. 431-435, 1972.

[9] F. D. Whisler, B. Acock, D. N. F. R. E. Baker, H. F. Hodges, J. R. Lambert, H. E. Lemmon and J. M. R. V. R. McKinion, "Crop simulation models in agronomic systems," *Advances in agronomy,* vol. 40, pp. 141-208, 1986.

[10] D. N. Baker, J. R. Lambert and J. M. McKinion, "GOSSYM: a simulator of cotton crop growth and yield," 1983.

[11] U. S. D. A., "United States: Cotton Production," [Online]. Available: https://ipad.fas.usda.gov/countrysummary/images/US/cropprod/USA_Cotton.png. [Accessed 1 August 2023].

[12] F. Chollet and others, *Keras,* Github, 2015.

[13] N. R. Prasad, N. R. Patel and A. Danodia, "Crop yield prediction in cotton for regional level using random forest approach," *Spatial Information Research,* vol. 29, pp. 195-206, 21.

[14] L. H. Nguyen, J. Zhu, Z. Lin, H. Du, Z. Yang and W. a. J. F. Guo, "Spatial-temporal multi-task learning for within-field cotton yield prediction," in *Advances in Knowledge Discovery and Data Mining: 23rd Pacific-Asia Conference, PAKDD 2019,*, Macau, China, 2019.

[15] S. Leo, M. D. Antoni Migliorati and P. R. Grace, "Predicting within-field cotton yields using publicly available datasets and machine learning," *Agronomy Journal,* vol. 113, no. 2, pp. 1150-1163, 2021.